\documentclass{article}

\usepackage{microtype}
\usepackage{graphicx}
\usepackage{subfigure}
\usepackage{booktabs} 

\usepackage{hyperref}

\usepackage[accepted]{icml2023}

\usepackage{amsmath}
\usepackage{amssymb}
\usepackage{mathtools}
\usepackage{amsthm}

\usepackage[capitalize,noabbrev]{cleveref}

\theoremstyle{plain}

\theoremstyle{definition}

\theoremstyle{remark}

\usepackage[textsize=tiny]{todonotes}

\icmltitlerunning{Deceptive Alignment Monitoring}

\begin{document}

\twocolumn[
\icmltitle{Deceptive Alignment Monitoring}



\icmlsetsymbol{equal}{*}
\begin{icmlauthorlist}
\icmlauthor{Andres Carranza}{equal,cs}
\icmlauthor{Dhruv Pai}{equal,cs}
\icmlauthor{Rylan Schaeffer}{equal,cs}
\icmlauthor{Arnuv  Tandon}{equal,cs}
\icmlauthor{Sanmi Koyejo}{cs}
\end{icmlauthorlist}

\icmlaffiliation{cs}{Computer Science, Stanford University}

\icmlcorrespondingauthor{Rylan Schaeffer}{rschaef@cs.stanford.edu}

\icmlkeywords{Machine Learning, ICML}

\vskip 0.3in
]



\printAffiliationsAndNotice{\icmlEqualContribution} 

\begin{abstract}
As the capabilities of large machine learning models continue to grow, and as the autonomy afforded to such models continues to expand, the spectre of a new adversary looms: \textit{the models themselves}.
The threat that a model might behave in a seemingly reasonable manner, while secretly and subtly modifying its behavior for ulterior reasons is often referred to as deceptive alignment in the AI Safety \& Alignment communities. Consequently, we call this new direction \textit{Deceptive Alignment Monitoring}.
In this work, we identify emerging directions in diverse machine learning subfields that we believe will become increasingly important and intertwined in the near future for deceptive alignment monitoring, and we argue that advances in these fields present both long-term challenges and new research opportunities.
We conclude by advocating for greater involvement by the adversarial machine learning community in these emerging directions.
\end{abstract}
\section{Introduction}
\label{sec:intro}

Machine learning models are growing increasingly general-purpose while simultaneously being granted increasingly more autonomy. The combination of greater capabilities and greater freedom in choosing when and how to exercise those capabilities raises the spectre that models themselves may behave adversarially to human interests \cite{hubinger2021risks, hendrycks2021unsolved, ngo2023alignment}. In the AI Safety and Alignment communities, this threat is often referred to as deceptive alignment, and large-scale collaborations spanning leading machine learning organizations and institutions are focusing more on how to measure problematic capabilities of large-scale models via benchmarks \cite{pan2023rewards,shevlane2023model}.

In this Blue Sky submission, we aim to elaborate on the different possible threats that large models themselves pose when viewed as potential adversaries, and argue that new methods are urgently needed. These methods must go beyond evaluation on static benchmarks to active, dynamic monitoring of models at multiple stages in a model's life-cycle: dataset creation and curation, training and learning, network representations and circuit mechanisms.

\section{Deceptive Alignment Monitoring}
\label{sec:dad}

\subsection{Data Creation \& Curation}

The use of foundation models to generate and/or curate their own training data has recently emerged in both language and vision.
In language generation, Self-Instruct used one language model to bootstrap training data for fine-tuning a different language model \cite{wang2022self}, and similar methods have quickly become commonplace, e.g., \cite{peng2023instruction, sun2023principle, xu2023wizardlm}.
In curation, SemDeDup uses an high-performing language model to prune a dataset for training future language models \cite{abbas2023semdedup}, and DoReMi learns to weight data differently to improve performance when learning \cite{xie2023doremi}. In vision, \cite{li2023internet} propose endowing a model with the ability to learn to construct its own training dataset in a self-supervised manner by trawling through the internet. Such methods will continue to improve and proliferate.

These methods are relevant to deceptive alignment because if a highly capable, autonomous model does wish to change its behavior, creating or curating training data are straightforward mechanisms by which it can achieve such a goal. Currently, such methods have a high degree of human-in-the-loop, but as model scale and complexity increase, humans will play a vanishing role. To the best of our knowledge, we lack methods for monitoring the training data that our models generate or curate to check that the training data has not been selected in pursuit of some ulterior goal. This threat is spiritually similar to recent work demonstrating that web-scale datasets can be poisoned \cite{carlini2023poisoning}; however, in the setting studied by \citet{carlini2023poisoning}, the dataset is known, fixed and constructed by humans, whereas in this setting, the dataset is unknown, dynamic and constructed by models in a largely unsupervised manner.

\subsection{Training, Editing \& Learning}

Turning now from data to learning, there are two emerging trends that pose new challenges: using one model to train another model, and directly editing the parameters of a model to change its behavior.
The first trend appears commonly when using reward or preference models to fine-tune a language model, such as in Reinforcement Learning from Human Feedback \cite{ouyang2022training, bai2022training}, Reinforcement Learning from AI Feedback \cite{bai2022constitutional}, Reinforcement Learning for Feedback \cite{akyurek2023rl4f} and many more, e.g., image synthesis \cite{lee2023aligning}.
The second trend includes several methods for model editing, i.e., directly editing the model's parameters to change its behavior on specific tasks or inputs, e.g., \citet{meng2022locating, mitchell2022memory, meng2022mass, hartvigsen2022aging, ravfogel2022linear, jason2023detecting}. In some methods, models are directly trained to learn to modify their own parameters, e.g., \citet{irie2022modern, kirsch2022self,kirsch2022eliminating}.

Why does this matter for deceptive alignment?
If models can have ulterior motives and act on said motives, then enabling models to train each other or to edit themselves directly posses a significant threat, since a model can give itself or give another model capabilities that humans would rather the model(s) not possess.
To the best of our knowledge, we lack techniques to audit what learning signals one model transmits to another model, and we also lack techniques to understand whether model editing methods can themselves be poisoned; for a concrete example, it might be possible that by selective choosing a small number of seemingly innocent edits (e.g., ``US President: Donald Trump $\rightarrow$ Joe Biden" and ``World Chess Champion: Magnus Carlsen $\rightarrow$ Ding Liren"), the combination of edits induces an unpredictable and undesirable change in model behavior\footnote{However, we note that sometimes seemingly unpredictable changes are indeed predictable, e.g., \citet{schaeffer2023emergent}.}.



\subsection{Representations \& Mechanisms}

A key step towards addressing the identified challenges will be monitoring the representations and mechanisms of large-scale models for ``permissible" behavior \cite{christiano2022elk, christiano2022mechanisms}; here, ``permissible" behavior is loosely understood since the concept itself is in desperate need of sharper understanding. Loosely, if we think of a network as some map $f: \mathcal{X} \rightarrow \mathcal{Y}$, previous work has studied adversarial inputs coming from an unusual input distribution $\Tilde{p}(x)$ or backdoored models with unusual conditional distributions $p(y|x)$. We are therefore concerned not with unusual input data, nor unusual conditional distributions, but whether the network's outputs are produced for the ``normal reasons". As \citet{christiano2022elk} explains, we are looking for outliers in terms of how $f$ handles perfectly normal data.

The reason why this matters is that while inputs, outputs, and inputs-to-output maps may all appear perfectly ordinary, highly capable models may conceal some ulterior motive or goal. In order to detect and counter this threat, it is imperative to develop interpretability methods capable of understanding both activations and circuit mechanisms of models and detecting abnormal variations in models' processing. This is important because two models generating identical outputs when provided with the same inputs may compute their outputs for different ``reasons".

Because the exact threat is unknown and likely dynamic, there is an urgent need to develop unsupervised methods for mechanistic anomaly detection that scale well. The next step is to develop an unsupervised methodology that can differentiate between normal model behaviors—where conclusions are reached for ``appropriate reasons"—and anomalous model behaviors—where conclusions are derived for erroneous, potentially harmful ``reasons". This task is particularly challenging since it requires the identification of patterns in the propagation of information through a model that is indicative of correct reasoning without relying on explicitly supervisory signals.

To achieve this, we propose leveraging techniques related to unsupervised anomaly detection to capture deviations from typical model behaviors. By comparing a model's processing across various inputs and outputs, it may be possible to identify patterns that consistently align with desired and appropriate behavior. We hypothesize that these patterns could manifest at three different levels of analysis within a model. Firstly, at the individual layer, a comprehensive analysis of activation distributions in the high-dimensional activation space could provide valuable insights into the model's processing. Secondly, at the layer-to-layer activation level, investigating how high-dimensional modes propagate, transform and evolve through the layers of a model can also offer an understanding of normal and abnormal processing. Thirdly, at the circuit level, identifying subgraphs within the network that correspond to specific transformations on features relevant to out-of-domain generalization might also prove powerful; however, knowing how to usefully define probabilistic distribution over activations, activations' propagations and circuit mechanisms for anomaly detection are, to the best of our knowledge, open questions. For possible approaches, see \citet{carranza2023facade}.

 \section{Outlook}
 \label{sec:conclusion}

The human-model interpretability quest can be modeled as an adversarial game, whereby deceptively aligned models subvert interpretability tools in favor of capabilities. More capable models are increasingly threatening, and to maintain scalable oversight we advocate development of novel tools for deceptive alignment monitoring. 
\clearpage

\bibliographystyle{icml2023}
\bibliography{references}

\end{document}